\documentclass[final]{cvpr}

\usepackage{times}
\usepackage{epsfig}
\usepackage{graphicx}
\usepackage{amsmath}
\usepackage{amssymb}
\usepackage{verbatimbox}

\usepackage[pagebackref=true,breaklinks=true,colorlinks,bookmarks=false]{hyperref}

\def\cvprPaperID{8706} 
\def\confYear{CVPR 2021}

\begin{document}

\title{Self-supervised Low Light Image Enhancement and Denoising}

\author{Yu~Zhang  \\
\and
Xiaoguang~Di\\
\and
Bin~Zhang\\
\and 
Qingyan~Li\\
\and 
Shiyu~Yan\\
\and 
Chunhui~Wang\\
 Department of Electronic Science and technology, Harbin Institute of Technology\\
{\tt\small zhangyuhit2@hit.edu.cn}
}

\maketitle

\begin{abstract}This paper proposes a self-supervised low light image enhancement method based on deep learning, which can improve the image contrast and reduce noise at the same time to avoid the blur caused by pre-/post-denoising. The method contains two deep sub-networks, an Image Contrast Enhancement Network (ICE-Net) and a Re-Enhancement and Denoising Network (RED-Net). The ICE-Net takes the low light image as input and produces a contrast enhanced image. The RED-Net takes the result of ICE-Net and the low light image as input, and can re-enhance the low light image and denoise at the same time. Both of the networks can be trained with low light images only, which is achieved by a Maximum Entropy based Retinex (ME-Retinex) model and an assumption that noises are independently distributed. In the ME-Retinex model, a new constraint on the reflectance image is introduced that the maximum channel of the reflectance image conforms to the maximum channel of the low light image and its entropy should be the largest, which converts the decomposition of reflectance and illumination in Retinex model to a non-ill-conditioned problem and allows the ICE-Net to be trained with a self-supervised way. The loss functions of RED-Net are carefully formulated to separate the noises and details during training, and they are based on the idea that, if noises are independently distributed,  after the processing of smoothing filters (\eg mean filter), the gradient of the noise part should be smaller than the gradient of the detail part. It can be proved qualitatively and quantitatively through experiments that the proposed method is efficient.
\end{abstract}
\begin{figure}[htbp]
\centering
\includegraphics [width=\linewidth]{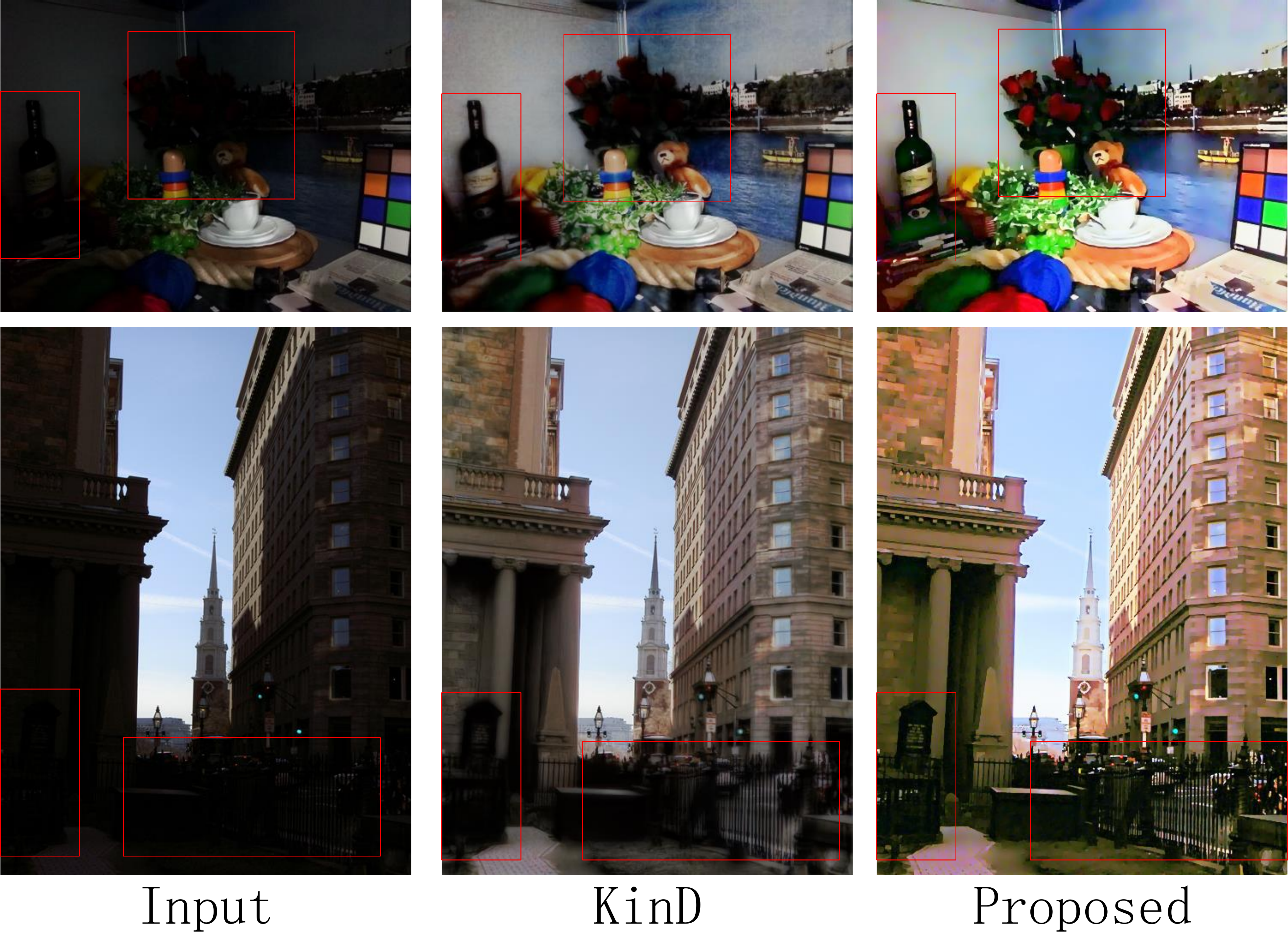}
\caption{Visual comparison with supervised low light image enhancement method KinD \cite{zhang2019kindling}. The proposed method can well improve the contrast and brightness of the images and at same time reduce noise, and sharpen the edges. (Best viewed on high-resolution displays with zoom-in)}
\label{kind_me}
\end{figure}
\section{Introduction}
\par Images captured in low light conditions always suffer from low contrast, low brightness and serious noise, and so on. Low light image enhancement method is used to solve those problems before high-level computer vision tasks, but there are few methods to deal with these problems well at the same time. Recently, various deep learning based algorithms have achieved surprising results in some image processing and computer vision tasks, such as object detection \cite{liu2016ssd}, \cite{ren2015faster}, \cite{redmon2018yolov3},\cite{he2017mask}, image segmentation \cite{he2017mask}, \cite{long2015fully}, \cite{badrinarayanan2017segnet}, etc. One most important reason for the rapid development of deep learning in these tasks is that we can obtain a large number of data sets with clear and unambiguous labels. In these tasks, although the construction of the data set requires some cost, it is still acceptable, also on the Internet, a large number of open-source data sets can be found for these tasks to support the training of the network. 
\par However, in low-level image processing tasks such as low light image enhancement, image dehazing, and image restoration, etc., it is difficult to obtain a large number of true input/label image pairs. As for low light image enhancement task, in the previous work, some supervised solutions such as synthesizing low light images \cite{Lore2015LLNet}, using images with different exposure time \cite{chen2018learning}, and so on, have achieved good visual effects, especially in noise reducing. Even though, there are still two problems with those methods. One is how to ensure that the pre-trained network can be used for images collected from different devices, different scenes, and different lighting conditions rather than building new training data set(\eg \cite{zhang2019kindling} failed to remove noises in the background in Fig. \ref{kind_me}). The other is how to determine whether the normal light image used for supervision is the best, there can be lots of normal light images for one low light image. Usually, the data sets with paired low/high light images are built with artificial adjustment, which will cost lots of time and energy, and also we cannot make sure the normal light images can complete the training task very well.
\par In this paper, to overcome the problems in previous works, we proposed a self-supervised low light image enhancement framework to realize image contrast enhancement and denoising at the same time. Similar to previous works\cite{Lore2015LLNet}, two networks are used to achieve contrast enhancement and denoising, respectively. However, different from the previous work, the two networks, i.e. Image Contrast Enhancement Network (ICE-Net) and the Re-Enhancement and Denoising Network (RED-Net), are trained self-supervised. And the RED-Net is designed to reduce noises through re-enhancing the contrast of low light images, which can reduce the loss of information caused by pre-/post-processing with existing denoising methods. 
\par For the training of ICE-Net, a Maximum Entropy based Retinex (ME-Retinex) model was proposed. Different from previous Retinex models which only assume that illumination is smooth, in the ME-Retinex model, we introduce a new constraint on the reflectance image that the maximum channel of the reflectance image conforms to the maximum channel of the low light image and its entropy should be largest. With a constraint on reflectance, we can directly control the image enhancement level and convert the ill-conditioned decomposition of reflectance and illumination into a non-ill-posed problem. 
\par For the training of RED-net, we adopt the assumption that the noises conform to the Poisson distribution and in different pixels they are independent. With this assumption, the gradients of noises should be smaller than the gradients of details after the processing of smoothing filters, and most of them are even close to zero. Then it is possible for us to separate most noises and details in reflectance by treating those gradients calculated from smoothed reflectance as weights. At the same time, considering that our task is to enhance the image contrast, and edges with higher gradients often have higher contrast, so the loss functions of RED-Net can be designed to make the gradients of details and edges higher to achieve better contrast enhancement, and it is different from the previous works which only try to preserve edges. Based on those ideas, some self-supervised loss functions are formulated and their effectiveness are proved through experiments. 
\par The loss function in this paper can complete self-supervised training, which means that we can directly solve image enhancement task for one specific image, whether by CNN(Convolutional Neural Network) or analytical methods. However, more training data in CNN often leads to better results(\eg Fig.\ref{fig_CNN}), and most of time CNN spends less processing time than analytical methods. The proposed method is independent of the way acquiring low light images, and the training process is completely self-supervised, so the method proposed in this paper has good generalization ability, even if the pre-trained network is not well enough in a new environment, retraining or fine-tuning it without building paired/unpaired normal light images data set is possible for the network.
Our contributions can be summarized as:
\begin{itemize}
\item We proposed a framework for enhancing low light images, which can enhance the image contrast and reduce noise at the same time. Through the close coupling of the two, we can reduce the loss of information in image enhancement tasks (\eg blur caused by commonly used post-denoising).
\item We proposed an Image Contrast Enhancement Network(ICE-Net) and a Re-Enhancement and Denoising Network(RED-Net), and both of them can be trained by self-supervision, which gets rid of the dependence on paired or unpaired images. Also, the RED-Net proposed in this paper can be combined with other Retinex or HSV based image enhancement methods to achieve re-enhancement and noise suppression, even AHE(Adaptive Histogram Equalization \cite{pizer1987adaptive}) which produces heavy noises, and this is helpful for many previous studies on contrast enhancement.  
\item We compare the proposed method with several state-of-the-art methods via some comprehensive experiments. And the results are measured by objective indexes and visual quality. All results consistently proved the effectiveness of the proposed method.
\end{itemize}
\section{Related works}
\textbf{Low light image enhancement.} Directly adjusting the contrast of the low light image is probably the most intuitive and easy way to realize image enhancement, such as Histogram Equalization(HE), and other improved methods based on HE \cite{pisano1998contrast, wang1999image, naik2003hue, celik2011contextual, lee2013contrast}. Although those improved methods are proposed to achieve noise suppression, hue preserving, brightness preserving, and so on, there are still many problems in directly adjusting the contrast, such as, over- and under- enhancement, noise amplification, \etal. Gamma correction is another kind of mapping manner, which is also a frequently used method for low light image enhancement. Although it can promise a well image brightness, and stretch the contrast in low or high areas, it still can not avoid noise amplification and most of the time, its result highly depends on the Gamma value which is chosen artificially. 
\par Retinex is a widely used model for low light imagencement in recently years. According to Retinex theory, an image can be decomposed into reflectance and illumination. The early works SSR \cite{jobson1997properties} and MSR \cite{jobson1997multiscale} treat the reflectance as the final enhancement result. However, since the decomposition is a ill-posed problem and without enough constraints on the reflectance, the enhanced image often have unreal phenomena such as the over-enhancement and whitening. Also, it is hard for those methods to reduce noise. In recent works, illumination are enhanced after the decomposition and the final enhanced image is obtained by recombining the enhanced illumination and reflectance. However, the enhanced image may still have noise and an extra post-denoising procedure have to be preformed\cite{guo2016lime,fu2016weighted}, which will produce blur in details. \cite{jed2018} introduced a joint low-light enhancement and denoising method, which can achieve denoising and enhancement simultaneously. \cite{jed22018} further improved the method through considering a noise map compared with the conventional Retinex model. Although those methods are proposed to have promising results, most of them need multiple iterations for decomposition which will cost lots of time. Meanwhile, as there is not any method to automatically manipulate the illumination, the enhanced image may not have a proper contrast and usually need careful parameter tuning. 
\par Recently, the amazing performance of deep learning also inspired some promising works in low light image enhancement, including supervised works \cite{Lore2015LLNet,wei2018deep,zhang2019kindling,fu2020learning} and unsupervised works \cite{jiang2019enlightengan,zeroReference}. Most of the early works based on supervised learning train the networks with synthetic data sets, such as \cite{Lore2015LLNet, li2018lightennet, yang2016enhancement, shen2017msr}, etc. Although the data obtained by these works seems to be dark and noisy, they are still different from a natural one. Chen et al. \cite{chen2018learning} introduce a dataset which contains real raw low light images and corresponding raw high light images for training. As there can be lots of reference images for one input low light image, they introduce an amplification ratio in the network to achieve correspondence between the input and reference. This method can well solve the problem of noise and color distortion, however, the ratio must be chosen by user during test which limit the widely use of this method, and there maybe some over-/under-enhanced areas in the image with only one ratio. \cite{wei2018deep} introduced a dataset named LOL which contains real paird low and high images. And it introduced the Retinex model into the training process to connect the reflection images of the input and reference, and proposed to denoise on reflectance with BM3D\cite{dabov2007image}. However, it will still cause blur or remain noisy, and it's hard to find a balance between the two. \cite{zhang2019kindling} added a subnet called restoration-net to achieve denoising on reflectance, and provided an extra brightness ratio to control the illuminantion. However, during the test, it still need to manually adjust the ratio parameter to obtain better enhancement results. Although these methods use real low light data for training, due to the lack of constraint on the contrast of the enhanced image, it can not avoid the problem of over-enhancement (saturation) or under-enhancement in the enhanced image, even with artificial adjustment of parameters. In the unsupervised works, \cite{jiang2019enlightengan} proposed a GAN-based method which can be trained with unpaired data, but it cannot control the enhancement results. \cite{zeroReference} proposed a zero-reference low light image enhancement method, which can be trained without any paired or unpaired data. However, it did not provide any noise removal methods. 
\par\textbf{Image denoising}
Many denoising methods have been proposed over the past few decades, including conventional methods \cite{dabov2007image,Weighted2014} and learning based methods \cite{beyond2017denoise, Liu2018denoise, Agostinelli2013Adaptive}. However, those denoising methods are not specially designed for the low light image enhancement task. No matter pre-/post-processing with those method will caused details loss, and the learning based method may even invalid for different kind of noise distribution. \cite{zhang2020better} proposed an denoising method for low light image enhancement, however, it need to be trained with paired low/high light image data. Recently, \cite{noise2void} proposed a unsupervised denoising method named N2V, which can be trained with the noisy image only, however, in our tests, it will still cause blur even re-trained with the enhanced image.   
\begin{figure}[htbp]
\centering
\includegraphics [width=0.9\linewidth]{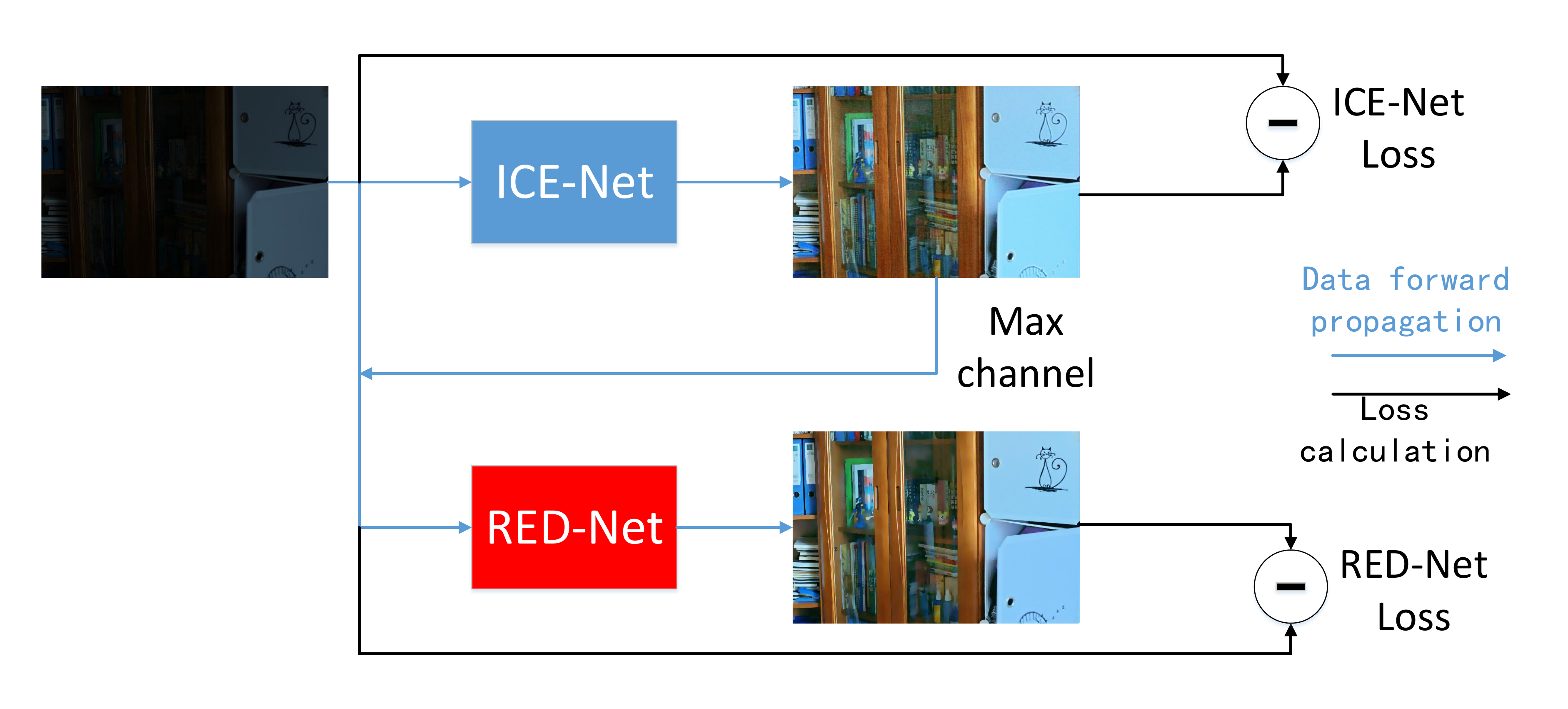}
\caption{ The structure of the proposed method, RED-Net take low light images and the max channel of the output of ICE-Net as input.  }
\label{Fig:structure}
\end{figure}

\section{Proposed Model}
\par The proposed method aims at achieving low light image enhancement and denoising without any artificial adjustment in the case of only having low light images. For example, when the camera enters a new low condition environment, the pre-trained networks may not works for different distribution, and the only data we can get are low light images. In order to achieve automatically contrast enhancement, we proposed a Maximum Entropy based Retinex model and an self-supervised ICE-Net to take advantage of multiple images. As for the denoising, we proposed an self-supervised RED-Net which is specially designed for the low lihgt image enhancement task.  Through the combination of re-enhancement and denoising, the RED-Net can save more details during denoising process. The structure of the proposed method is shown in Fig. \ref{Fig:structure}. 
\subsection{ME-Retinex model and ICE-Net}
\par Recently, a lot of low-light image enhancement works are based on the following Retinex model:
\begin{equation}
\label{eqn1}
\mathbf{S}=\mathbf{R}\circ \mathbf{I}
\end{equation}
\noindent where $\mathbf{S}$ and $\mathbf{I}$ represent the captured image and the illumination image respectively, $\mathbf{R}$ represents the reflectance and some works treat it as the desired enhanced image, $\circ$ represents the element-wise multiplication. Most of recent works assume that the three color channels of the image have the same illumination in order to simplify the model \cite{wei2018deep},\cite{zhang2019kindling}, and the maximum value of the three color channels is generally used as the initial estimate of the illumination map \cite{guo2016lime}. It has been proved that image enhancement methods based on this simplified Retinex model are equivalent to directly control contrast on V channel in HSV color space and remain the H and S channels unchanged \cite{zhang2020better}. However, there are still some differences between those two types of methods. Methods which directly control the image contrast usually stretch the contrast of some areas and compress the contrast of other areas, and the compressed areas will lose details(over-/under-enhancement can be treated as details missing). For example, HE will merge the smaller bins, and Gamma correction will compress the contrast of bright areas, both of those will cause the loss of details. And methods based on Retinex usually do not contain the constraints on the contrast of target enhanced image (whether $\mathbf{R}$ or $\mathbf{R} \circ \mathbf{I}^{\gamma}$ ), which produce the uncertain of results. 
\par If we transform the HE or Gamma correction to the Retinex model to explain, the enhanced image $\mathbf{R}$ is obtained through $\mathbf{S}/\mathbf{I}$ without the constraint that illumination $\mathbf{I} $ is smooth, then the missing details will be retained in the illumination $\mathbf{I}$. It can be considered that, if a rich texture area $\mathbf{S}$ is divided by a smooth $\mathbf{I}$, the details will be in $\mathbf{R}$ which can avoid the loss of details in HE or Gamma correction. Then we can combine the method of directly controlling the contrast with the Retinex model to take advantage of the both. And the combination will suppress the noise through the assumption that illumination is smooth, compared with the method of direct control contrast.
\par Typically, a Retinex based method can be expressed as follows:
\begin{equation}
\label{eqn2}
\underset{\mathbf{R},\mathbf{I}}{min} l_{rcon}+\lambda_{1}l_{\mathbf{R}}+\lambda_{2}l_{\mathbf{I}}
\end{equation}
Where, $l_{rcon}$, $l_{\mathbf{R}}$ and $l_{\mathbf{I}}$ represent reconstruction loss, reflectance loss and illumination loss, respectively. $\lambda_{1}$ and $\lambda_{2}$ are weight parameters. 
The reconstruction loss $l_{rcon}$ can be expressed as: 
\begin{equation}
\label{eqn3}
l_{rcon}=\left \| \mathbf{S}-\mathbf{R}\circ \mathbf{I} \right \|_{1}
\end{equation}
Where $||\bullet||_{1}$ represents the $L_{1}$ norm, we use the $L_{1}$ norm to constrain all the losses, and do not compare the impact of $L_{1}$, $L_{2}$, SSIM and other loss functions on low level image processing tasks, since there are already some related studies such as \cite{zhao2016loss}.
\par In this paper, we choose the HE method to form a Maximum Entropy based Retinex model, then the reflectance loss is formulated as:
\begin{equation}
\label{eqn4}
l_{\mathbf{R}}=\left \| \underset{c\in{r,g,b}}{max} \mathbf{R}^{c} - F(\underset{c\in{r,g,b}}{max}\mathbf{S}^{c}) \right \|_{1} + \lambda \left \|\triangledown \mathbf{R} \right \|_{1}
\end{equation}
where $F(x)$ means the histogram equalization operator to image $x$. $\lambda$ is weight parameters, $\triangledown$ means gradient operator. This first term of this loss function means that maximum channel of the reflectance should conform to the maximum channel of the low light image and has the maximum entropy, which can be considered as directly control the contrast of enhanced image. The second term is a commonly used smoothing term to suppress noise, but usually it is hard to distinguish image details and noises well. 

\par For the illumination loss, we adopt the structure-aware smoothness loss proposed in \cite{wei2018deep}:
\begin{equation}
\label{eqn5}
l_{\mathbf{I}}=\left \| \triangledown \mathbf{I}\circ exp\left ( -\lambda_{3} |\triangledown \mathbf{R}|  \right ) \right \|_{1}
\end{equation}
\par It is proposed that Equation \ref{eqn5} can make the illumination loss aware of the image structure in \cite{wei2018deep}. And this loss means that the original TV function  $\left \| \triangledown \mathbf{I} \right \|_{1}$ is weighted with the gradient of reflectance.
\par
\begin{figure}[htbp]
\centering
\includegraphics [width=3.5in]{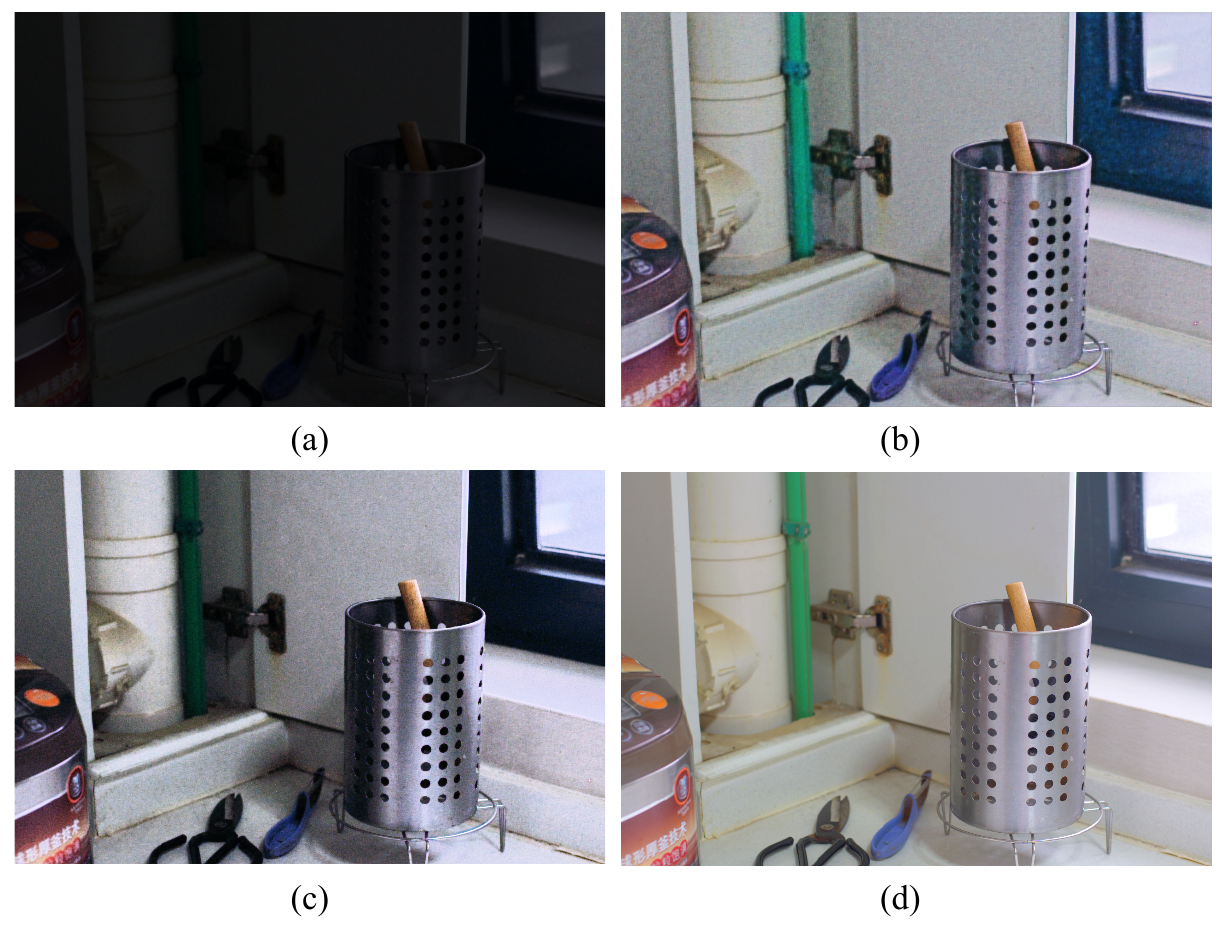}
\caption{The results by ICE-Net with different training data. (a) Input. (b) The network was trained with multiple data. SSIM:0.6743, PSNR:23.4716, NIQE:3.9140
  (c) The network was trained with (a) only. SSIM:0.4858, PSNR:15.5112, NIQE:4.9367  (d) Reference}
\label{fig_CNN}
\end{figure}

 For Equations \ref{eqn2}-\ref{eqn5}, we introduce an ICE-Net to solve this optimization problem. Then there will be a question, why introduce CNN to do that? As the ideal image can be obtained from minimizing the total loss, then one could just run this optimization directly on the $\mathbf{R}$ and $\mathbf{I}$ for a single image $\mathbf{S}$ and the introduction of CNN does not seem to be necessary. However, most of the optimization process need multiple iterations which will bring time consumption problems, and with more constraints, the solution will be more complicated. And at the same time, HE is a global enhancement method, which will inevitably lead to the problem of too bright or too dark in some local areas. By introducing CNN and training on multiple images, this problem can be avoided, as shown in Fig. \ref{fig_CNN}. This is because under the HE constraint, the same local area in different images will be enhanced to different degrees, and CNN will be trained to find the median value with $L_{1}$ regularization instead of becoming over bright or over dark. In addition, the loss function and ICE-Net is designed to learn how to get an appropriate enhancement in contrast and brightness, so we did not make any special design on denoising and the RED-Net designed in next sub-section can well achieve denoising.
\subsection{RED-Net}
\par After the processing of ICE-Net, although the contrast of the image has been improved, there are still some noises in the image. Inspired by \cite{zhang2020better} which introduce a Conditional Re-Enhancement Network(CRE-Net) to denoise for low light image enhancement tasks, we further propose a self-supervised RED-Net to re-enhance the low light image and denoise at the same time. In this part, we still build the loss function based on Equation \ref{eqn2}, however, every sub-loss function has been modified. For the reconstruction loss $l{}'_{rcon}$ and reflectance loss $l{}'_{\mathbf{R}}$ in RED-Net, we both adopt the assumption that the noise conforms to the Poisson distribution\cite{poisson} which is more in line with the real low light image noises. In order to distinguish from the variables in ICE-net, we added the superscript ${}'$ for the variables in RED-Net and the reconstruction loss can be expressed as:

\begin{equation}
\label{eqn6}
l{}'_{rcon}= \mathbf{R{}'}\circ \mathbf{I{}'}-\mathbf{S} \circ log\left(\mathbf{R{}'}\circ \mathbf{I{}'}\right) 
\end{equation}
where $\mathbf{R{}'}$ and $\mathbf{I{}'}$ represents the reflectance and illumination produced by RED-Net,respectively, and $\mathbf{R{}'}$ is also target enhanced image of whole proposed method. 

\par
\begin{figure}[htbp]
\centering
\includegraphics [width=\linewidth]{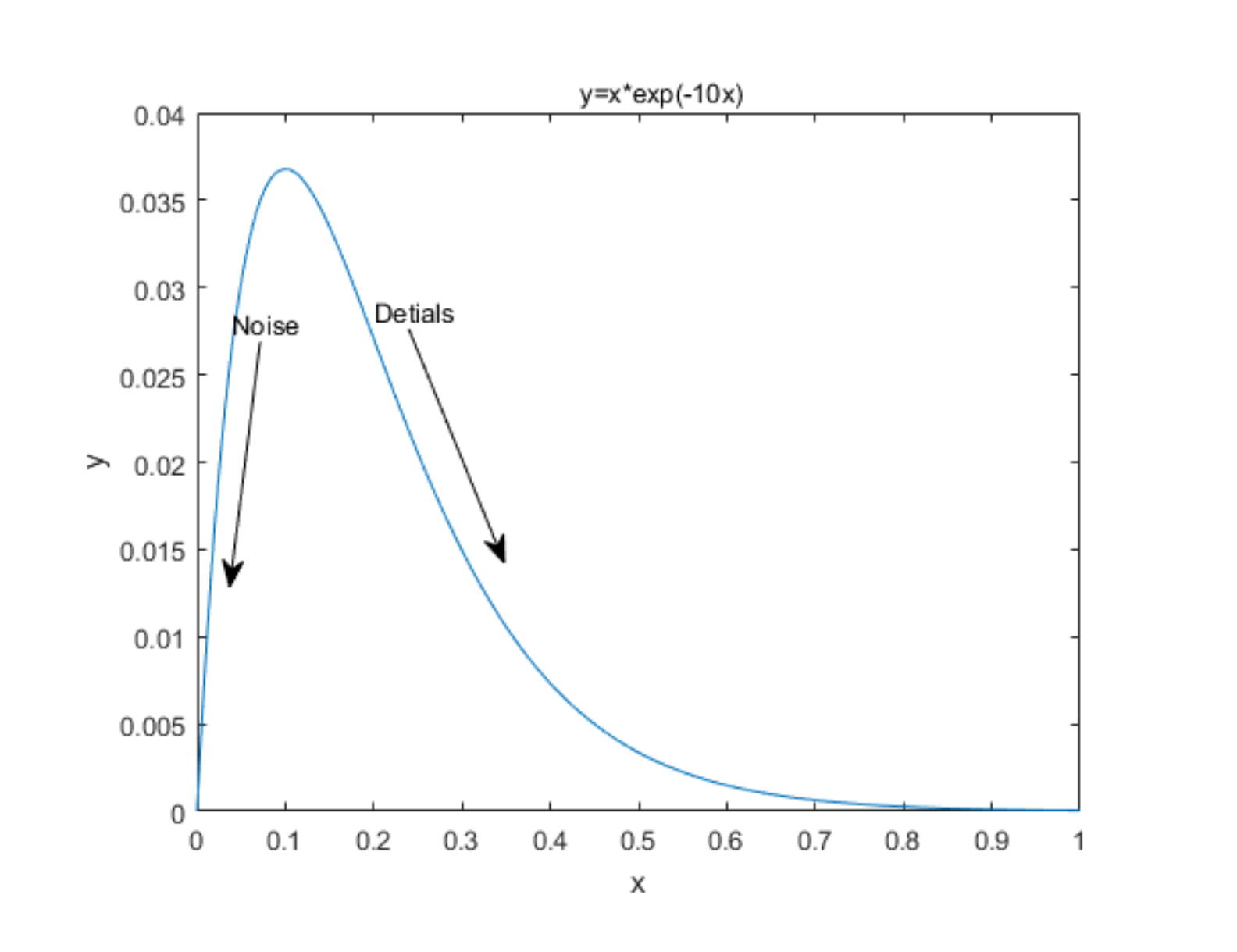}
\caption{The curve of $y=x*exp\left(-\lambda x\right)$}
\label{fig_loss_cure}
\end{figure}

\begin{figure*}[htbp]
\centering
\includegraphics [width=\linewidth]{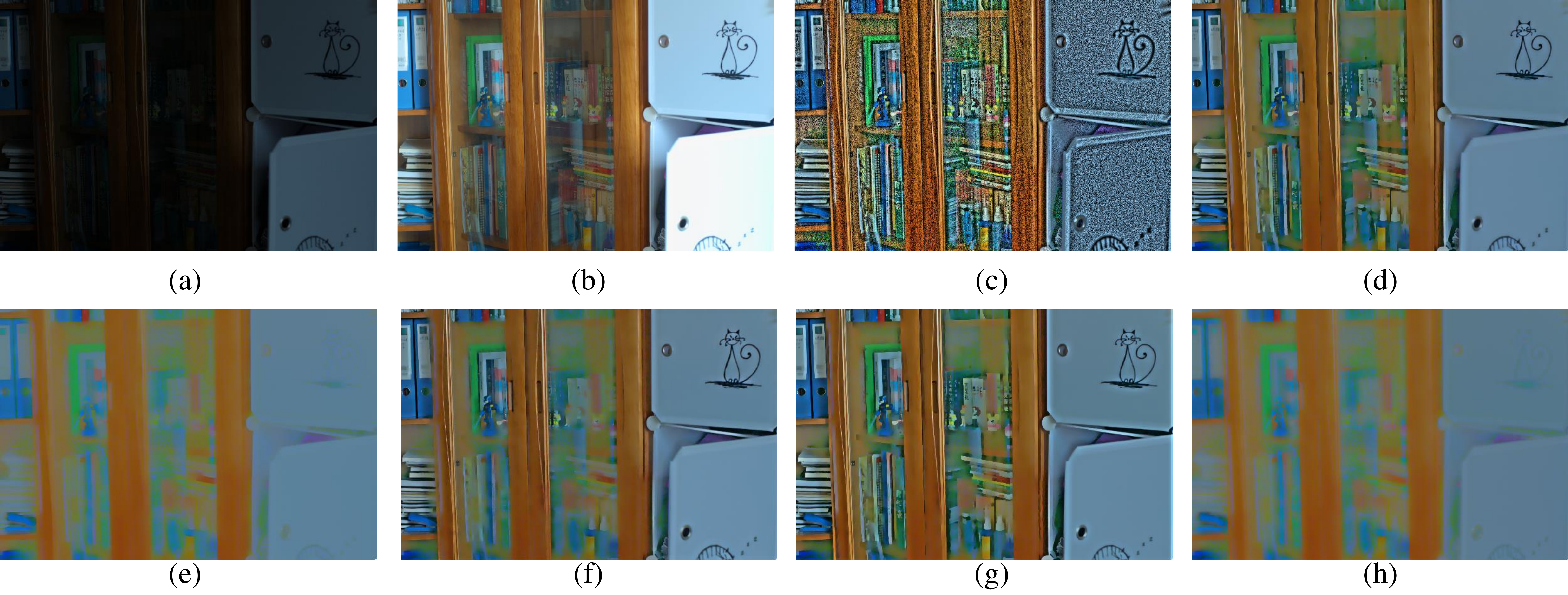}
\caption{The results generated with different loss in RED-Net. (a) Input low. (b) Reference high. (c) AHE\cite{pizer1987adaptive}. (d) AHE \& RED-Net. (e) w/o $\mathbf{W}$. (f) w/o $\mathbf{W_I}$ and $\mathbf{W_R}$ . (g) w/o $exp \left( -\lambda_{3}\mathbf{W_I} \circ N\left( |\triangledown \mathbf{I{}'}|  \right)\right)$. (h) w/o $exp\left( -\lambda_{3}  \mathbf{W_R} \circ N\left( |\triangledown \mathbf{R{}'} |\right) \right)$. }
\label{fig_loss}
\end{figure*}

 We argue that an image can be divided into different components, including noise, flat area, details and structure information, and there is no clear dividing line between details and structure information. There are many methods to remove noise, but the key to the problem is how to separate details and structure from noise, then to preserve or even strengthen those details and structure during denoising. In order to make reflectance less noise, and preserve rich details and sharp edges, we design the reflectance loss as follows:
\begin{align}
\label{eqn7}
l{}'_{\mathbf{R}}=& \underset{c\in{r,g,b}}{max} \mathbf{R{}'}^{c} - \underset{c\in{r,g,b}}{max} \mathbf{R}^{c} \circ log(\underset{c\in{r,g,b}}{max} \mathbf{R{}'}^{c})
\nonumber\\
{+}\:&\lambda \left\| \mathbf{W} \circ N(| \triangledown \mathbf{R{}'}|)\circ exp(-\lambda_{3} \mathbf{W} \circ N(|\triangledown \mathbf{R{}'}|)) \right\|_{1}
\end{align}
\noindent where $N(x)$ and $|x|$ represent the local normalization on $x$ and absolute value of $x$, respectively. $\mathbf{R}$ and $\mathbf{R{}'}$ represent the output reflectance of the ICE-Net and RED-Net, respectively. $\mathbf{W}$ represents weights, which can be calculated as follows:
\begin{equation}
\label{eqn8}
\mathbf{W}= N \left( |\triangledown \left(G \left( \mathbf{R{}'} \right) \right)|\right)
\end{equation}
where, $G(x)$ represents smooth filter on $x$(The mean filter are used in proposed method). The graph made by the second term $x*exp\left(-\lambda x\right) $ is shown in Fig. \ref{fig_loss_cure}. Intuitively, after smoothed, there are still gradients in the details and structure, even they are smaller than before. But the noise and smooth areas may have no gradients or have much smaller gradients. Then we can use the gradients of those smoothed images as the weight. As it can be seen in Fig. \ref{fig_loss_cure}, when the loss function is in the form of $x*exp\left(-\lambda x\right)$, small $x$ will become smaller, and high $x$ will become higher during training. And through the local normalization, the details and structure are more likely to fall on the right and make them more sharper during training. As shown in Fig. \ref{fig_loss}, noise are well removed and the details are preserved.
\par Typically, illuminations are usually expected to retain only structural information and ignore detailed information. Therefore we can adopt a design similar to reflection loss, and the illumination loss can be expressed as follows:

\begin{align}
\label{eqn9}
l{}'_{\mathbf{I}}=& || \mathbf{W_I} \circ N\left( |\triangledown \mathbf{I{}'}| \right)\circ exp \left( -\lambda_{3}\mathbf{W_I} \circ N\left(|\triangledown \mathbf{I{}'}|\right)  \right) \nonumber\\
\circ & exp\left( -\lambda_{3}  \mathbf{W_R} \circ N\left(|\triangledown \mathbf{R{}'} | \right)\right)  ||_{1}
\end{align}

\noindent where $\mathbf{W_I}$ and $\mathbf{W_R}$ represents weights, which can be calculated as follows:

\begin{equation}
\label{eqn10}
\mathbf{W_I}= N \left( |G  \left( \triangledown \mathbf{I{}'} \right)|\right)
\end{equation}

\begin{equation}
\label{eqn11}
\mathbf{W_R}= N \left( | G  \left( \triangledown \mathbf{R{}'} \right)|\right)
\end{equation}
\noindent where $G(x)$ and $N(x)$ still represent smooth filter and local normalization on $x$, respectively. Different from $\mathbf{W}$ in Equation \ref{eqn8}, the order of gradient operation $\triangledown$ and smoothing operation $G$ are switched. It can be considered that,
for noise and details, the mean value of the gradient in local area should be small, which is quite different for the structure. For example, texts in a white paper may have opposite gradients in a local area, which makes the mean gradient close to zero. Then with the $\mathbf{W_I}$ and the special design loss form $x*exp\left(-\lambda x\right)$, we can separate the noise and details from the structure in illumination, and make the structure edge sharper during training. Also we preserve $exp\left( -\lambda_{3}  \mathbf{W_R} \circ N\left( |\triangledown \mathbf{R{}'} |\right) \right)$ and introduce the weight $\mathbf{W_R}$ to ensure the consistency of the structural information of the reflectance and the illumination. It should be noted that all weight items $\mathbf{W}$, $\mathbf{W_R}$, $\mathbf{W_I}$, do not participate in the back propagation process during training.

\section{Experiments}
\par  We use the LOL database \cite{wei2018deep} which contains 500 low/normal light image pairs, 485 of which are used for training and each image size is $400*600$. Note that during the training process, we only use natural low light images without any synthetic data and normal light images. During the training process, our batch size is set to 16 and the patch size is set to 48 * 48. We use Adam stochastic optimization \cite{kingma2014adam} to train the network and the update rate is set to 0.001. The training and testing of the network are completed on a Nvidia GTX 2080Ti GPU and Inter Core i9-9900K CPU, and the code is based on the tensorflow framework.
\par To evaluate the performance of the proposed method on enhancing low-light images, we quantitatively and visually compare our method with some low light image enhancement methods, including LIME\cite{guo2016lime}, RRM \cite{jed22018}, Retinex-Net\cite{wei2018deep}, KinD \cite{zhang2019kindling}, and also we collected some data from other data sets for testing. 
\par Three metrics are adopted for quantitative comparison, which are Peak Signal-to-Noise Ratio(PSNR), Structural SIMilarity(SSIM) \cite{SSIM2004}, and NIQE \cite{mittal2012making}. NIQE is a non-reference image quality assessment method, which can evaluate the naturalness of the image and a lower value indicates better quality. While, PSNR and SSIM are referenced image quality assessment methods, which indicate the noise level and the structure similarity between the result and the reference, respectively.

\subsection{Ablation Study}
\par In this part, to prove the necessity of introducing the CNN and the effectiveness of each component of the proposed method, we have made two ablation studies. 
\par\textbf{Contribution of ICE-Net.} This ablation study is to answer the question that why not just optimize the loss function to get the result, like other variational based Retinex models\cite{kimmel2003variational,park2017low}, if the network can be trained in a self-supervised way. As mentioned in Sec. 3.1, the CNN based method ICE-Net is introduced in our proposed method to avoid the problems caused by HE through training with multiple data. Considering it is difficult to directly solve Equation \ref{eqn2} through variational methods under proposed loss functions in this paper, we use a CNN trained with only a single low light image instead, and the result can be considered as a solution to the Equation \ref{eqn2}. 
\par In Fig. \ref{fig_CNN}, we present the results of our ICE-Net trained with one single low light image and with multiple images, and the results show that training with multiple data has a better enhancement effect in contrast and brightness. It can be seen in Fig.\ref{fig_CNN}(c), optimization on a single low light image cannot avoid the the problems of under-enhancement or over-enhancement (\eg green pipes and metal hinges), which is caused by HE. However, in Fig. \ref{fig_CNN}(b), training with multiple images, every local area of the enhanced image have a more proper brightness. Also it can be seen through objective indexes, training with multiple data shows better results in PSNR, SSIM and NIQE, which means that the enhanced result with multiple training data has less noise and seems more like reference and is more in line with natural images.  

\begin{figure}[htbp]
\centering
\includegraphics [width=\linewidth]{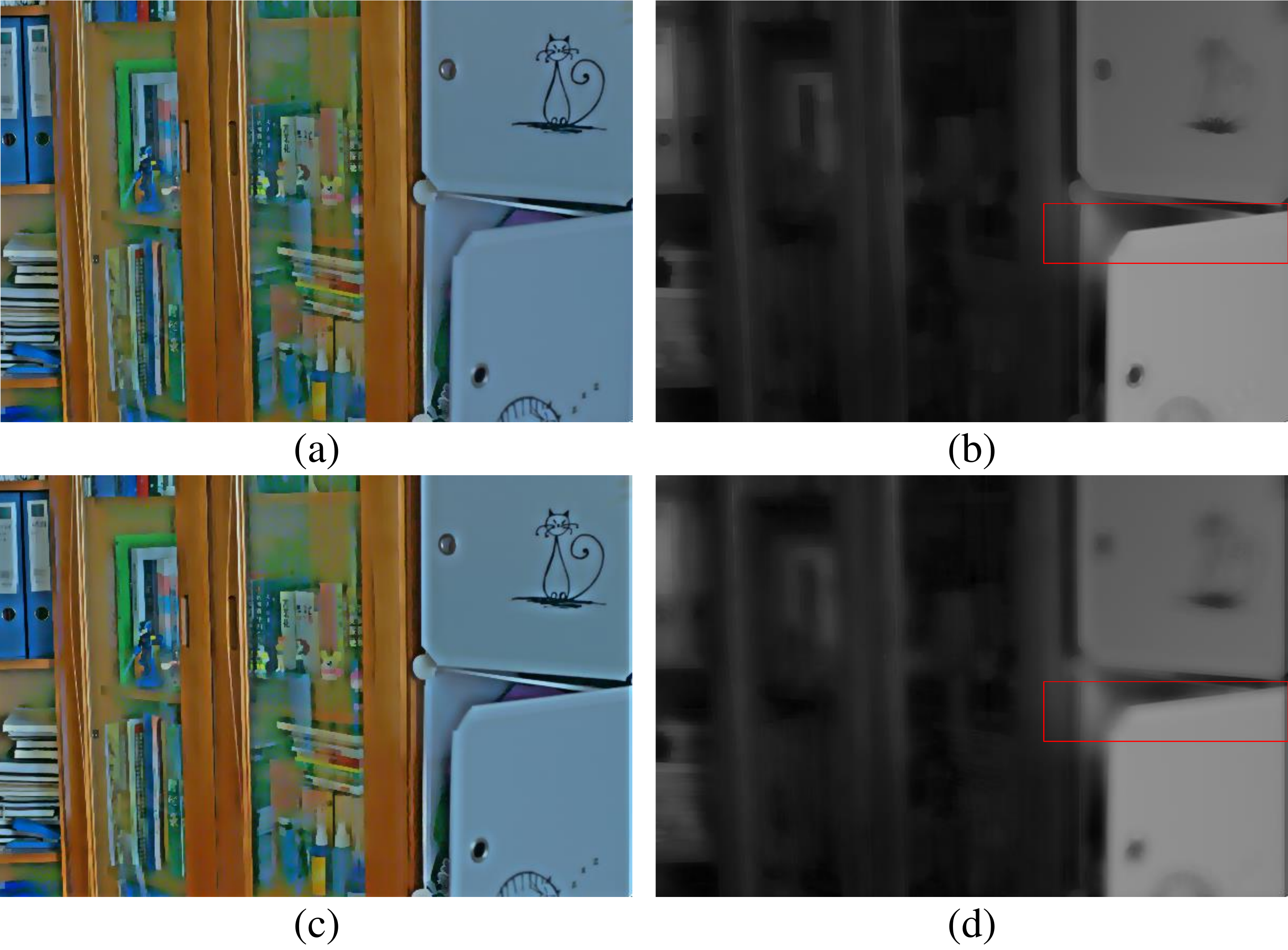}
\caption{Visual comparison with different loss in RED-Net.(a)-(b) and (c)-(d) are training with or without $exp \left( -\lambda_{3}\mathbf{W_I} \circ N\left( |\triangledown \mathbf{I{}'}|  \right)  \right)$, respectively. Please zoom in to see the details}
\label{FIg:illumination}
\end{figure}

\begin{figure*}[htbp]
\centering
\includegraphics [width=\linewidth]{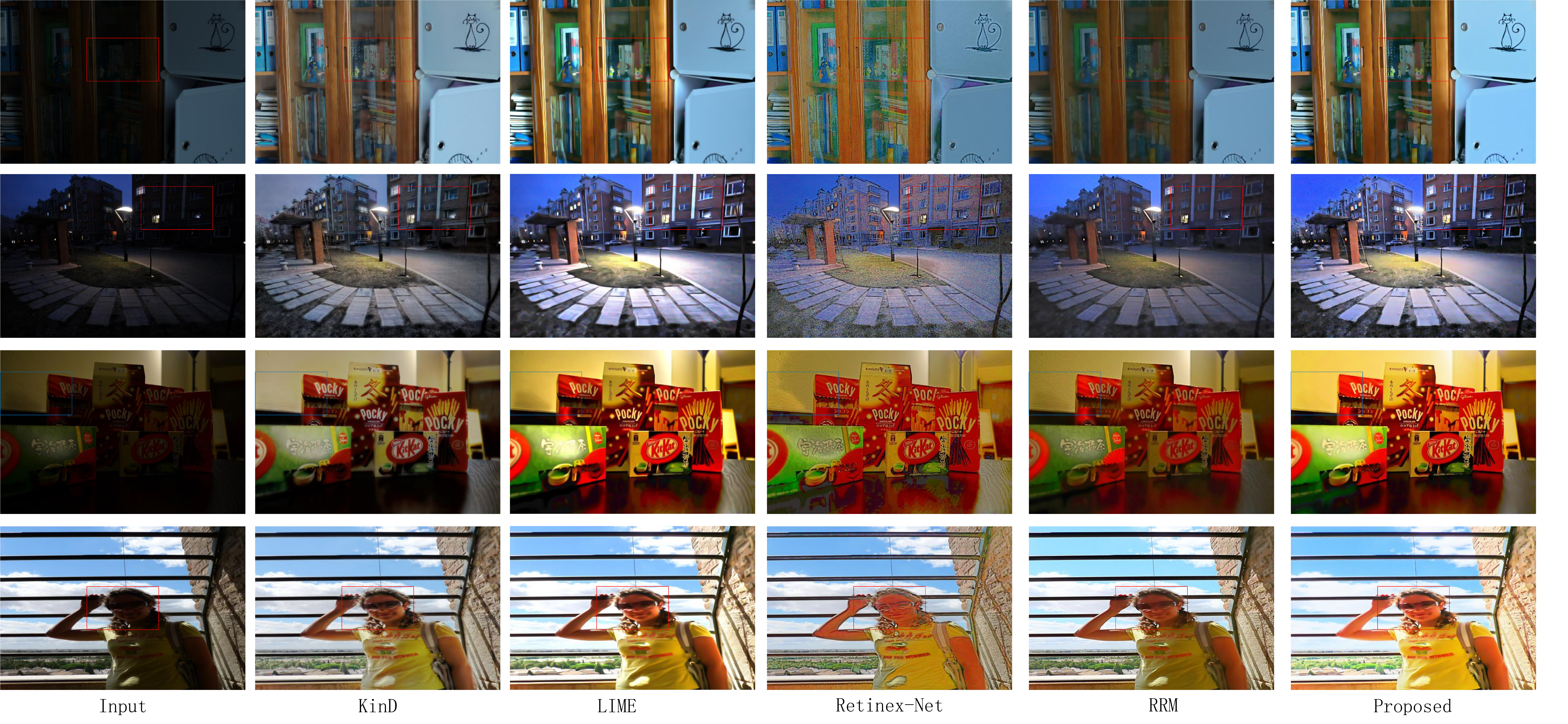}
\caption{Visual comparison with other state-of-the-art methods, each row comes from a different data set and each column comes from a different method. From left to right are Input, KinD\cite{zhang2019kindling}, LIME\cite{guo2016lime}, Retinex-Net\cite{wei2018deep}, RRM\cite{li2018structure},and proposed method in this paper. Please zoom in to see the details}
\label{fig_SOTA}
\end{figure*}
\par\textbf{Contribution of Each loss in RED-Net.} We present the results of RED-Net trained by different losses and weights in Fig. \ref{fig_loss}. In order to better illustrate the importance of each part of the loss function, we take the AHE\cite{pizer1987adaptive} which produces serious noise during processing as the contrast enhancement method(\eg Fig. \ref{fig_loss}(c)), and study the re-enhancement and denoising effects under different loss functions. We use the complete loss function as the baseline (Completely contains Equation \ref{eqn6} to \ref{eqn11}), and study the influence of removing different weight terms, including
\begin{itemize}
\item without $\mathbf{W}$ in Equation \ref{eqn7}, (Fig. \ref{fig_loss} (e))
\item without $\mathbf{W_I}$ and $\mathbf{W_R}$ in Equation \ref{eqn9}, (Fig. \ref{fig_loss} (f))
\item without $exp \left( -\lambda_{3}\mathbf{W_I} \circ N\left(|\triangledown \mathbf{I{}'}| \right) \right)$ in Equation \ref{eqn9}, (Fig. \ref{fig_loss} (g))
\item without $exp\left( -\lambda_{3}  \mathbf{W_R} \circ N\left(|\triangledown \mathbf{R{}'} |\right) \right)$ in Equation \ref{eqn9}, (Fig. \ref{fig_loss}(h))
\end{itemize}
\par As shown in Fig.\ref{fig_loss} (d), with all the proposed loss functions, the RED-Net can obviously reduce noise and at the same time preserve details. When we simply remove $\mathbf{W}$ (\eg Fig.\ref{fig_loss} (e)), only the obvious struture are preserved, that proves the importance and effectiveness of separating the noise and details through $\mathbf{W}$.  And when we remove $exp\left( -\lambda_{3}  \mathbf{W_R} \circ |\triangledown \mathbf{R{}'} | \right)$ (\eg Fig. \ref{fig_loss} (h)), the details are lost and some obvious edges are slightly blurred. And when we remove $\mathbf{W_I}$ and $\mathbf{W_R}$ which are designed to smooth noise and details and preserve structure in illumination, some details in reflectance are blurred(\eg Fig.\ref{fig_loss} (f)), which proves the importance of smoothing in the part of noises and details in illumination, and also proves the effectiveness of our design. 
\par In Fig. \ref{fig_loss} (g) and (d), it seems that the third kind of loss in this ablation study which removes $exp \left( -\lambda_{3}\mathbf{W_I} \circ |\triangledown \mathbf{I{}'}|  \right)$ in illumination loss does not affect the result of reflectance. However, it can be seen in Fig.\ref{FIg:illumination} (b) and (d), the edges (\eg edge in red rectangles) in illumination are blurred under the third kind of loss, which may caused halo effect in reflectance, and a well illumination can help a lot in future work too(\eg avoiding over-enhancement).  We also study the case that Poisson distribution is not used. However, with AHE \cite{pizer1987adaptive}, the output of the RED-Net is totally unacceptable, and even the structural cannot be saved.
\subsection{Comparison with State-of-the-Arts}
\par In this subsection we compare the performance of the proposed method with current state-of-the-art methods through qualitative and quantitative experiments. And during these experiments, we used not only the LOL dataset, but also some standard datasets collected from previous works, including LIME\cite{guo2016lime}(10 images), MF(10 images), and VV \footnote{https://sites.google.com/site/vonikakis/datasets/challenging-dataset-for-enhancement} (23 images).
\par We have compared the combination of ICE-Net and RED-Net with some previous methods which can achieve contrast enhancement and denoising, including LIME\cite{guo2016lime} which has a denoising post-processing, RRM\cite{li2018structure} which can jointly enhance contrast and denoise, Retinex-Net\cite{wei2018deep} which is trained through supervised ways and denoise with BM3D\cite{dabov2007image} in reflectance, KinD\cite{zhang2019kindling} which is trained through supervised ways in contrast enhancement and denoising, and the code is download from the author's homepage and parameters are set as recommended in those paper. The results are shown in Fig. \ref{fig_SOTA} and Table \ref{tab:NIQE} and \ref{tab:SSIM}.
\par Fig.\ref{fig_SOTA} shows the qualitative evaluation results, it can be seen that, compared with the LIME and Retinex-net which denoise   on the reflectance with BM3D, the two-stage method proposed in this paper which first enhances, and then re-enhances and denoises can keep a better balance between denoising and detail preserving, and the method is even comparable to the supervised method KinD (\eg Books in the bookcase processed by LIME are blurred and processed by Retinex-Net still have serious noise, and both our method and KinD can well preserve the texts in the book). Also it can be seen in the second and last row of Fig. \ref{fig_SOTA}, our method is able to work under serious noise and non uniform illumination conditions(\eg. the face and arm in the shadow state are well enhanced). At the same time, since we assume that the difference between detail and noise is that noise is distributed independently, which is not always right, the rough wall was smoothed in the blue rectangle in the third row. (More detailed experiments, comparisons, network structure and parameters are included in the supplementary material. In our experiments, the impact of network structure is not significant, and RED-Net has the same network structure as in \cite{zhang2020better}, and the ICE-Net is similar to RED-Net, but has less layers than RED-Net).   
\par Table \ref{tab:NIQE} and \ref{tab:SSIM} show the quantitative evaluation results, it can be seen that, our method gets poor NIQE, and highest PSNR and middle SSIM, which means that after finally enhanced, the image processed by our method seems different from the natural image and reference, and noises in the images are well removed. This is due to that, during designing the ICE-Net and RED-Net, we mainly consider the automatic adaptation ability of the algorithm to the new environment and the removal of noise, and do not take any natural image prior into loss functions, especially in the RED-Net. And at the same time, in the process of enhancement, our goal is to enhance the details of each local area, which is quite different from the reference image obtained by adjusting the exposure time.
\begin{table}
	\centering
	\caption{NIQE scores on the each subset(LOL \cite{wei2018deep}, LIME \cite{guo2016lime}, MF \cite{fu2016fusion}, VV), and smaller NIQE indicate more in line with natural images.}
	\resizebox{0.48\textwidth}{!}
	{\addvbuffer[0pt 0pt]{
		\begin{tabular}{cccccc}
			\hline
			Dataset & LIME\cite{guo2016lime} & RRM\cite{li2018structure}  & Retinex-Net\cite{wei2018deep} &KinD\cite{zhang2019kindling} & Proposed \\ 
			\hline
			LIME \cite{guo2016lime} & 4.08  & 4.03 & 4.37 & \bf{3.59}  & 5.07 \\
			LOL \cite{wei2018deep}  & 3.95  & 3.95 & 9.06 & \bf{3.89}  & 4.33 \\
			MF \cite{fu2016fusion}  & 3.44  & 3.68 & 3.88 & \bf{3.31}  & 4.59 \\
			VV                      & 3.22  & 3.31 & 3.57 & \bf{2.91}  & 4.01 \\
			\hline
		\end{tabular}
		}
	}
	\label{tab:NIQE}
\end{table}
\begin{table}
	\centering
	
	\caption{SSIM amd PSNR scores on the LOL \cite{wei2018deep} data set, and higher SSIM and PSNR indicate more in line with reference and less noise, respectively.}
	\resizebox{0.48\textwidth}{!}
	{
	\addvbuffer[0pt 0pt]{
		\begin{tabular}{cccccc}
			\hline
			Dataset & LIME\cite{guo2016lime} & RRM\cite{li2018structure}  & Retinex-Net\cite{wei2018deep} &KinD\cite{zhang2019kindling} & Proposed \\ 
			\hline
			PSNR  & 17.22  & 13.88 & 16.82  & 17.64      &\bf{18.34} \\
			SSIM  & 0.60   & 0.66  & 0.57   & \bf{0.76}  & 0.65 \\
			\hline
		\end{tabular}
		}
	}
	\label{tab:SSIM}
\end{table}




\section{Conclusion and Future Work}
\par In this paper, aiming at automatically enhancing the low light images and denoising, we create a two-stage framework which enhances the image contrast first and then further re-enhances and denoises. And both of the networks in our method can be trained with a self-supervised way, which means that the proposed method can be used in real new unfamiliar environment and new device. The experimental results on various low light data sets show that our method is comparable with many state-of-the-arts methods on both visual effect and subjective metrics. Our future works will explore how to restore the color degradation, how to combine the RED-Net and the ICE-Net, and how to the combine the low light image enhancement and high-level tasks to further improve real-time performance.   

{\small
\bibliographystyle{ieee_fullname}
\bibliography{egbib}
}

\end{document}